\documentclass{article}
\usepackage{arxiv}
\usepackage[utf8]{inputenc} 
\usepackage[T1]{fontenc}    
\usepackage{hyperref}       
\usepackage{url}            
\usepackage{booktabs}       
\usepackage{amsfonts}       
\usepackage{nicefrac}       
\usepackage{microtype}      
\usepackage{lipsum}
\usepackage{tablefootnote}
\usepackage{graphicx} 
\usepackage{tabularx}
\usepackage{color,soul}
\usepackage{pbox}
\usepackage[thinlines]{easytable}

\title{An Iterative Scientific Machine Learning Approach for Discovery of Theories Underlying Physical Phenomena}

\author{
  Navid~Zobeiry \\
  Materials Science \& Engineering Department\\
  University of Washington\\
  \texttt{navidz@uw.edu} \\
   \And
 Keith D. ~Humfeld \\
  Boeing Research and Technology\\
  \texttt{Keith.D.Humfeld@boeing.com} \\
}

\begin{document}
\maketitle

\begin{abstract}
Form a pure mathematical point of view, common functional forms representing different physical phenomena can be defined. For example, rates of chemical reactions, diffusion and heat transfer are all governed by exponential-type expressions. If machine learning is used for physical problems, inferred from domain knowledge, original features can be transformed in such a way that the end expressions are highly aligned and correlated with the underlying physics. This should significantly reduce the training effort in terms of iterations, architecture and the number of required data points. We extend this by approaching a problem from an agnostic position and propose a systematic and iterative methodology to discover theories underlying physical phenomena. At first, commonly observed functional forms of theoretical expressions are used to transform original features before conducting correlation analysis to output. Using random combinations of highly correlated expressions, training of Neural Networks (NN) are performed. By comparing the rates of convergence or mean error in training, expressions describing the underlying physical problems can be discovered, leading to extracting explicit analytic equations. This approach was used in three blind demonstrations for different physical phenomena.
\end{abstract}

\keywords{Theory-Guided Machine Learning \and Scientific Machine Learning \and Physics-informed Deep Learning \and Neural Networks \and Knowledge Discovery}

\section{Introduction}
Machine Learning (ML) has the potential to significantly speed-up scientific discoveries and transform all branches of science and engineering \cite{lee2018basic}. To employ machine learning in physical problems, theory and domain knowledge can be leveraged to setup domain-aware models. This approach which is recently coined as Scientific Machine Learning (SciML) \cite{lee2018basic}, Theory-guided Machine learning (TGML) \cite{karpatne2017theory,wagner2016theory,zobeiry2019TGML1,zobeiry2019TGML2} or Physics-informed Deep Learning \cite{raissi2017physics1,raissi2017physics2}, offers many advantages over traditional black-box machine learning methods. This includes capability to train models with small and fragmented data (which is often the case in physical problems) \cite{karpatne2017theory}, setup interpretable models with physics-based architecture, and ensure well-posedness and physical consistency in solution \cite{lee2018basic}. A variety of approaches have been recently explored in the scientific literature to incorporate theory and domain knowledge in machine learning including imposing physical constraints or symmetries in models, enriching loss functions using physics-based correlations, and feature transformation based on governing physical laws \cite{wagstaff2001constrained,wang2017physics,karpatne2017physics,bird2007transport}.

Any physical phenomenon such as heat transfer, turbulence flow, wave propagation or radiation damping, can be described using a set of partial differential equations usually on the basis of conservation laws such as energy, mass or momentum \cite{feynman2011feynman}. In some specific cases, these equations can be solved to obtain closed-form solutions representing underlying correlations and physics of the problem (e.g. steady-state heat transfer) \cite{bergman2011fundamentals}. Generally speaking, one might also express equivalent solutions using phenomenological models that are consistent with the underlying theory but are obtained through empirical observations and measurements. 

Considering a specific physical phenomenon, instead of setting up partial differential equations or developing empirical solutions, one might employ machine leaning algorithms to train neural networks based on measurable inputs (i.e. features) and outputs (e.g. temperature). When the training is performed using transformed inputs that describe the mathematics underlying the physical phenomenon (e.g. a combination of the variables that is present in the mathematical expression) \cite{karpatne2017theory}, the neural network trains more quickly and the errors in predicting outputs are decreased. Given this understanding, it must be possible to identify combination of variables that are present in the underlying methodical expression: train a neural network with the input data augmented by an expression that is a combination of inputs, and the rate at which the each neural network trains should be a measure of how well the combined variable represents components of the mathematical expression. This can be done in the absence of any knowledge of the meanings of the variables and without any knowledge of the components of the mathematical expression (i.e. theory-agnostic). What we are therefore seeking to do is to introduce a new method for discovering theories underlying physical phenomena, by employing an iterative and physics-based feature transformation to find components of the mathematical expression of the physical problem. This is done on the basis of commonly observed functional forms of physical phenomena. To explore the applicability of the method, in a blind approach, three demonstrations from different physical domains are considered in this paper. Results showed that the proposed method is a promising tool to uncover underlying expressions, leading to constructing closed-form solutions.

\section{Distilling Functional Forms of Physical Phenomena}
Considering the immense body of theory in various fields of science and engineering, it becomes a daunting task to distill functional forms of available expressions and closed-form solutions. However, families of similar forms can be observed:

\begin{itemize}
\item Crosses: in many theoretical fields, such as heat transfer and fluid mechanics, specific unit-less crosses or non-dimensional numbers are shown to highly influence the physics of the problem. Not surprisingly, these numbers appear in many closed-form solutions \cite{bergman2011fundamentals}.  Mathematically speaking, often times these are nothing but crosses of original features. Prime example is the Reynolds number, \textit{Re}, in fluid mechanics that is used to identify fluid regimes (i.e. laminar or turbulence), and appears in countless closed-form solutions \cite{bird2007transport}:
\begin{equation} 
\label{EQ_1} 
Re=\frac{\mathrm{inertial}~\mathrm{forces}}{\mathrm{viscous}~\mathrm{forces}}=\frac{uL}{\nu } 
\end{equation} 
Here \textit{Re} is expressed as the cross of three original features: velocity, $u$, characteristic length, $L$, and kinematic viscosity, $\nu $.

\item
Power functions: This is also a very common functional form in physical problems. Often times, crosses and power functions appear together. In solid mechanics, for example, deformation of a simply supported beam, $\delta $, under a center load, $P$, is obtained on the basis of the Timoshenko beam theory \cite{lubliner2016introduction}:
\begin{equation} 
\label{EQ_2} 
\delta =\frac{PL}{4}\left(\frac{L^2}{12EI}+\frac{1}{kGA}\right) 
\end{equation} 
Here \textit{E} and \textit{G} are related to beam material properties, \textit{I, A} and \textit{L} are related to beam dimensions, and\textit{ k} is a constant (i.e. shear correction factor).\textit{ }

\item  
Exponential forms: This is a widely observed functional form in various fields including statistical mechanics, heat and mass transfer, solid mechanics and polymer engineering. Prime example is exponential decay such as Arrhenius equation that describes the dependency of rates of a chemical reactions, $k$, on temperature, \textit{T} \cite{ladd1998introduction}: 
\begin{equation} 
\label{EQ_3} 
k=Ae^{\frac{-E_a}{RT}} 
\end{equation} 
In this equation, $A$ and $E_a$ are empirically measured constants and \textit{R} is the universal gas constant. Other examples include Newton's law of cooling in heat transfer \cite{bergman2011fundamentals} and Boltzmann's law in statistical mechanics \cite{feynman2011feynman}.

\item  
Others: Other common types of functional forms include hyperbolic, trigonometric and logarithmic. These are listed in Table~\ref{Table_1}. 
\end{itemize}

\begin{table}[!htb]
\setlength\extrarowheight{10pt}
\caption{Summary of common functional forms of physical phenomena}
\centering
\begin{tabular}{lll}
\toprule
	Category & Examples \\ \hline
	Crosses & Reynold number in fluid mechanics \cite{bird2007transport}: $Re=\frac{uL}{\nu }$ \\
	Power & Beam bending in solid mechanics \cite{lubliner2016introduction}: $\delta =\frac{PL}{4}\left(\frac{L^2}{12EI}+\frac{1}{kGA}\right)$ \\
	Exponential & Exponential decay in chemical reactions \cite{ladd1998introduction}: $\ k=Ae^{\frac{-E_a}{RT}}$ \\
	Hyperbolic & Fin convective heat transfer rate \cite{bergman2011fundamentals}: $Q'=M\frac{{\mathrm{sinh} mL\ }+\frac{h}{mk}{\mathrm{cosh} mL\ }}{{\mathrm{cosh} mL\ }+\frac{h}{mk}{\mathrm{sinh} mL\ }}$ \\
	Trigonometric & Capillary adhesion of two flat plates \cite{de2013capillarity}: $F=\pi R^2\frac{2\gamma {\mathrm{cos} {\theta }_E\ }}{H}$ \\
	Logarithmic & Isothermal entropy change of an ideal gas \cite{bent1965second}: $\Delta S=R\times ln\left(\frac{V_2}{V_1}\right)$ \\ 
\bottomrule
\end{tabular}
\label{Table_1}
\end{table}

\section{Method}
We sought to explore the idea described above through a single-blind study in which one researcher created data corresponding to a mathematical expression describing a physical phenomenon (i.e. a known theory result), flatten that data to remove all description of what the variables indicated (i.e. substituted generic variable names and removed all units) and provided the set of inputs and outputs to the second researcher. The second researcher trained a series of neural networks utilizing the inputs and outputs and one additional combination of the variables as a unique input to each neural network. It was quickly determined that this approach would not be successful in any reasonable amount of time. The computational effort required to train a neural network repeatedly, in order to determine the rate at which the training occurs, and to do this for each possible combination of input variables was far too much. A lower cost method of screening potential combinations of variables was needed.

That method was a correlation analysis. Using Python and sklearn library, a first program was written to iteratively develop functional expressions of input variables. A second program took these functional expressions of input variables and determined the degree to which the outputs were correlated to each of those functional expressions. The most correlated functional expressions were then used individually to augment the inputs into a neural network which was then trained for an increasing size of training data in order to determine the rate at which the neural network was trained with that additional input. These steps are described below:

\begin{enumerate}
\item  For a given problem, initially the original \textit{n }features ($X_1\ to\ X_n$)  were mathematically transformed to create a library of base functions, $Z$, as listed in Table~\ref{Table_2}:

\begin{table}[!htb]
\setlength\extrarowheight{10pt}
\caption{List of base functions, \textit{Z}, created from original features}
\centering
\begin{tabular}{lll}
\toprule
	Category & Initial Base Functions (\textit{Z}) \\ \hline
	Original & $X_i\ $  \\  
	Cross/Power & $X^a_i\mathrm{,\ \ }X^{a_i}_iX^{a_j}_j$,  $X^{a_i}_iX^{a_j}_jX^{a_k}_k$     \\ 
	Exponential & $e^{cX^{a_i}_i}$,  $e^{cX^{a_i}_iX^{a_j}_j}$ \\
	Hyperbolic & ${\mathrm{sinh} \left(cX^a_i\right)\ },\ {\mathrm{cosh} \left(cX^a_i\right)\ },\ {\mathrm{tanh} \left(cX^a_i\right)\ },\ {\mathrm{coth} \left(cX^a_i\right)\ }$ \\ 
	Trigonometric & $sin(cX_i$) ,$cos(cX_i$) ,$tan({cX}_i$) ,$\ cot({cX}_i$) ,$sec(cX_i$) ,$\ csc(cX_i$)\\
	 & ${sin}^2(cX_i$) ,${cos}^2(cX_i$) ,${tan}^2({cX}_i$) ,$\ {cot}^2({cX}_i$) ,${sec}^2(cX_i$) ,$\ {csc}^2(cX_i$) \\
\bottomrule
\end{tabular}
\[i,j,k\in \left[1\dots n\right], a\in \left[-3,\ -2.5,-2,-1.5,-1,-0.5,\ 0.5,\ 1,\ 1.5,2,\ 2.5,3\right], c\in \left[0.1,\ 0.25,\ 0.5,\ 1,\ 2,\ 5,\ 10\right]\] 
\label{Table_2}
\end{table}

\item
All base functions and outputs (i.e. vectors) were normalized using their respective mean, $Z_{mean}$, and standard deviation, $std$, values:
\begin{equation} \label{EQ_4} 
\overline{Z}=\frac{Z-Z_{mean}}{std} 
\end{equation} 

\item  
The covariance of each output vector and base function was calculated to obtain their correlations: 
\begin{equation} \label{EQ_5} 
cov\ \left(\overline{Z},\overline{Y}\right)=\frac{\sum{\left({\overline{Z}}_i\right)}\left({\overline{Y}}_i\right)}{N} 
\end{equation} 

\item  
For some cases, top 10-20 highest correlated base functions were used to calculate new base functions:
\begin{equation} \label{EQ_6} 
b_1{\overline{Z}}_1+{\overline{Z}}_2\ \ \ ,    {{\overline{Z}}_1\left(b_2+{\overline{Z}}_2\right)}^a 
\end{equation} 
In these functions, constants are obtained using linear regression: 
\begin{equation} \label{EQ_7} 
\frac{\partial }{\partial b_1}\sum{{\left(b_1{\overline{Z}}_{1i}+{\overline{Z}}_{2i}-{\overline{Y}}_i\right)}^2}=0\to b_1=\frac{\sum{{\overline{Z}}_{1i}\left({\overline{Y}}_i-{\overline{Z}}_{2i}\right)}}{\sum{{{\overline{Z}}_{1i}}^2}} 
\end{equation} 
For some cases, this process of creating new bases functions from previous set of functions was repeated couple of times. 

\item  
From the list of highest correlated base functions, random combination of functions (2 to 6) were selected as transformed inputs to train a neural network with Fully Connected (FC) layers to predict the output. A high-Level API, estimator, in TensorFlow was used for all training. For a given dataset, size of training dataset was incrementally increased in consecutive training to study the effect of dataset size. List of all hyper-parameters are given in Table~\ref{Table_3}.
\begin{table}[!htb]
\setlength\extrarowheight{10pt}
\caption{Hyper parameters for training of Neural Networks}
\centering
\begin{tabular}{lll}
\toprule
	Parameter & Value \\ \hline
	Neurons & 3 hidden layers, 5 nodes per layer  \\  
	Activation & $\mathrm{ReLU}$  \\ 
	Optimization & Proximal Adaptive Gradient \\
	Learning rate & $0.04$ \\ 
	Bath-size & 50 \\
	Steps & 25000 \\ 
\bottomrule
\end{tabular}
\label{Table_3}
\end{table}

\item  
After all training, base functions that were creating the biggest hits in the output (i.e. lowest value of error, RMSE) were chosen and the process was repeated to find all related base functions.
\end{enumerate}

\section{Demonstrations}

\subsection{Demonstration 1}
A set of 2000 data, comprising six inputs ($X_1\ \mathrm{to}\ X_6)$  and one output ($Y$) consistent with an unidentified theory, were provided by researcher one. To preserve the anonymity of the data, units for the inputs and outputs were not provided. 

\begin{figure}[!htb]
  \centering
  \includegraphics[width=5in]{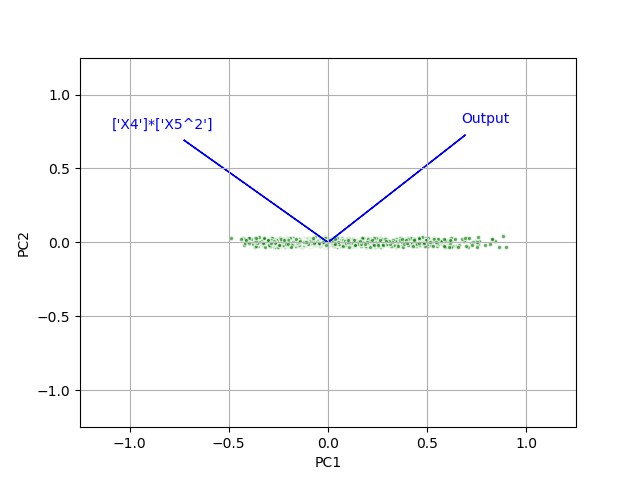}
  \caption{PCA analysis for a highly correlated function to output}
  \label{fig:image1}
\end{figure}

\begin{figure}[!htb]
  \centering
  \includegraphics[width=5in]{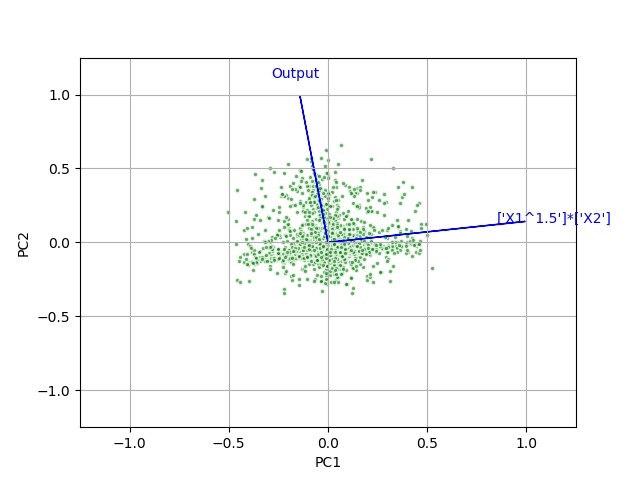}
  \caption{PCA analysis for a randomly correlated function to output}
  \label{fig:image2}
\end{figure}

An observation from the correlation analysis was that two of the original inputs, $X_2\ \mathrm{and}\ X_6$, were only randomly correlated with the output. Reducing the number of inputs from 6 to 4, greatly reduced the number of functions that would be built by the function generator, allowing to explore the iterative process more effectively. The correlation analysis was fast enough that large number of base functions could be analyzed quickly on a quad-core Intel i7 processor. From this first analysis, three functions showed the highest correlations to output: $X_4{X_5}^2$, $X_4{X_5}^{2.5}$, $X_4{X_5}^{1.5}$ with $X_4{X_5}^2$ as the top base function with 99.4\% correlation to output. In Figure \ref{fig:image1} and Figure \ref{fig:image2}, results from the Principle Component Analysis (PCA) for two different functions, one highly correlated, $X_4{X_5}^2$, and one randomly correlated to output, ${X_1}^{1.5}X_2$, are shown. From this first analysis, it was observed that results from correlation analysis was dominated by various derivations of $X_4\ \mathrm{and}\ X_5$ functions. To reduce the dependency of output, $Y$, to $Z=X_4{X_5}^2$ base function,  new output was created using the following equation:
\begin{equation} \label{EQ_8} 
Y_{new}=Y-\frac{\mathrm{max}\mathrm{}(Y)}{\mathrm{max}\mathrm{}(Z)}\ Z 
\end{equation} 
With this new output, PCA analysis was performed again and different sets of highly correlated functions were obtained. This process was repeated for the third time. Results from these analysis to identify top correlated functions in each step are shown in Table~\ref{Table_4}. 

\begin{table}[!htb]
\setlength\extrarowheight{10pt}
\caption{Iterative PCA analysis to find highly correlated functions}
\centering
\begin{tabular}{lll}
\toprule
	First PCA trial & Second PCA trial & Third PCA trial \\ \hline 
	$X_4{X_5}^2$ & $X_3$ $X_5$ & $X_1$ \\  
	$X_4{X_5}^{2.5}$ & $X_3{X_5}^{1.5}$ & \\ 
	$X_4{X_5}^{1.5}$ & $X_3{X_5}^{0.5}$ & \\
\bottomrule
\end{tabular}
\label{Table_4}
\end{table}

\begin{figure}[!htb]
  \centering
  \includegraphics{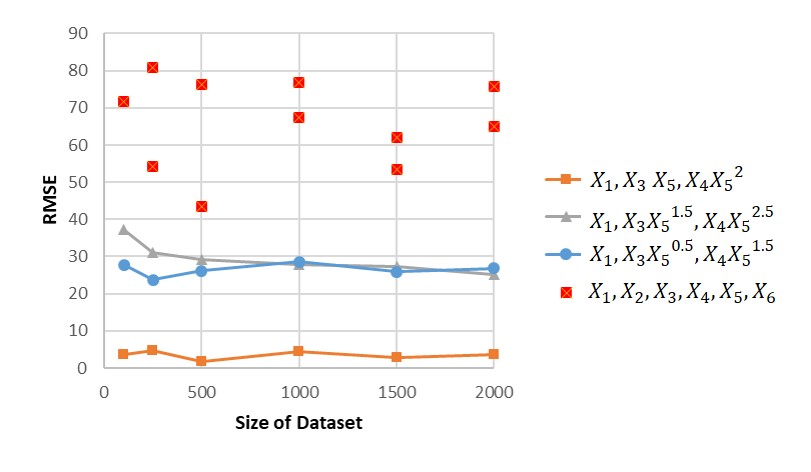}
  \caption{RMSE from several NN training using different combinations of input functions}
  \label{fig:image3}
\end{figure}

To establish a baseline, the recipient of the data trained a neural network using the six original provided inputs.  This model was trained several times using successively more data points ranging from 100 to 2000. The Root Square Mean Error (RSME) of the predictions of each trained neural network were recorded from each training. Random combinations of functions were then selected from the list in Table~\ref{Table_4} and NN training with different dataset sizes were performed again. RMSEs of several trainings for 4 combinations of inputs are compared in Figure \ref{fig:image3}. This shows that a combination of $X_1,{\ X}_{\mathrm{3}}~X_{\mathrm{5}},{\ X}_{\mathrm{4}}{X_{\mathrm{5}}}^{\mathrm{2}}$  base functions gives the lowest error in NN training.

Even though NN training was used in this demonstration, it was not greatly advantageous to do so. If the intention is to discover a combination of variables included in the mathematics that describes a physical phenomenon, finding a base function that is 99.4\% correlated to the output meets the objective without relying on neural networks. This was suspected, however, to be related to the simplicity of the mathematical expression in the first demonstration. The equation used for demonstration one describes the one-dimensional problem of motion under constant acceleration:
\begin{equation} \label{EQ_9} 
x=x_0+v_0t+\frac{1}{2}at^2 
\end{equation} 
Here, $x_0=X_1$ is initial position of a movable object, $v_0=X_3$ is the initial velocity of the object, $a=X_4$ is the constant acceleration and $t=X_5$ is time. The function generator achieved a correlation better than 99\% but not 100\% due to the limitations of the function generator itself. Later an iterative capability was built into the function generator it was able to produce a function that had precisely 100\% correlation to the data as a result of precisely matching the equation used without relying on NN training.

\section{ Demonstration 2}
Similar to previous demonstration, a dataset of 6 inputs ($X_1\ \mathrm{to}\ X_6)$ and 1 output ($Y$) with 2000 data points were provided by the first researcher. Based on the success of the function generator for demonstration one, it was expanded to include more functional forms of the input variables. Specifically, the following base function was added to the list in Table~\ref{Table_2}:
\begin{equation} \label{EQ_10} 
X^{a_i}_i{\left({cX}^{a_j}_j+X^{a_k}_k\right)}^n 
\end{equation} 
Using correlation analysis, the function generator created a ranked list of expressions all highly correlated to output. Top correlated functions are listed here: 
\begin{equation} \label{EQ_11} 
\frac{\sqrt{X_2}}{{\left(\frac{2\times 0.187}{\sqrt{X_3}}+\sqrt{X_1}\right)}^2}  , \frac{X_2}{{\left(\frac{2\times 0.187}{\sqrt{X_3}}+\sqrt{X_1}\right)}^2} , \frac{\sqrt{X_6}}{{\left(\frac{2\times 0.187}{\sqrt{X_3}}+\sqrt{X_1}\right)}^2} , \frac{1}{{\left(\frac{2\times 0.187}{\sqrt{X_3}}+\sqrt{X_1}\right)}^2} , \frac{\sqrt{X_4}}{{\left(\frac{2\times 0.187}{\sqrt{X_3}}+\sqrt{X_1}\right)}^2} 
\end{equation} 
This shows a common form in all the highly correlated functions. A second iteration of correlation analysis revealed an expression with 99.5\% correlation to the output:
\begin{equation} \label{EQ_12} 
z=\frac{\sqrt{X_2X_4X_6}}{{\left(\frac{2\times 0.187}{\sqrt{X_3}}+\sqrt{X_1}\right)}^2} 
\end{equation} 
The first researcher with knowledge of the mathematical expression used to generate the data, found that the solution did bear resemblance to the desired expression, but had some problems. In particular there was unit mismatch, and also a component of the equation missing from the suggested expression. The equation used for demonstration two was an expression for the \textit{Nusselt} number describing the heat transfer coefficient for forced convection over a circular cylinder. This equation was described by Churchill and Bernstein \cite{churchill1977correlating} as:
\begin{equation} \label{EQ_13} 
Nu=0.3+\frac{0.62{Re}^{{1}/{2}}{Pr}^{{1}/{3}}}{{\left[1+{\left({0.4}/{Pr}\right)}^{{2}/{3}}\right]}^{{1}/{4}}}{\left[1+{\left(\frac{Re}{282,000}\right)}^{{5}/{8}}\right]}^{{4}/{5}} 
\end{equation} 
In which following non-dimensional numbers are used:
\begin{equation} \label{EQ_14} 
Nu=\frac{hD}{k},  Re=\frac{DV\rho }{\mu }, Pr=\frac{\mu C_p}{k}
\end{equation} 
In above equations, $h=Y\ $is the heat transfer coefficient, $D=X_1$ is the cylinder diameter,   $k=X_2$ is fluid thermal conductivity, $V=X_{3\ }$is the fluid velocity,$\ \rho =X_4$ is the fluid density and $C_p=X_6\ $is the specific heat capacity. The best correlated solution from the function generator (Equation 12) was:
\begin{equation} \label{EQ_15} 
h=\frac{\sqrt{k\rho C_p}}{{\left(\frac{2\times 0.187}{\sqrt{V}}+\sqrt{D}\right)}^2} 
\end{equation} 
Which can be rewritten as:
\begin{equation} \label{EQ_16} 
Nu=\frac{\sqrt{\frac{D^2V^2{\rho }^2}{{\mu }^2}\frac{{\mu C}_p}{k}\frac{\mu }{\rho }}}{{\left(0.374+\sqrt{\frac{DV\rho }{\mu }}\sqrt{\frac{\mu }{\rho }}\right)}^2}=\frac{Re{Pr}^{{1}/{2}}\sqrt{\frac{\mu }{\rho }}}{{\left(0.374+{Re}^{{1}/{2}}\sqrt{\frac{\mu }{\rho }}\right)}^2} 
\end{equation} 
Unit mismatch is possible to prevent. Given that the intention of this research is to show a new method for discovering theories based on data, it is not necessary that researchers strip away the unit information from the data. Unit information can be used to establish a set of constraints within the function generator such that unit mismatch will not occur. Evaluation of the second demonstration was not repeated with constraints on the function generator based on units, but this approach was carried forward to the later demonstrations.

The equation used in the second demonstration was an empirical equation determined as a fit to experimental data. This likely makes the equation inappropriate for inclusion in this study. The data generated for the second demonstration, as a simulated experiment, precisely matches the empirical equation which is only an approximate fit to actual experimental data. It is very possible that the empirical equation used in the second demonstration includes terms to help it match the data well, but that are not actually representative of the underlying physical phenomenon. Rather than attempting to improve the results for the second demonstration, given its potential inapplicability to this research, the research moved to a third demonstration.

\section{ Demonstration 3}
For the third demonstration, the researchers included knowledge of the units of the variables. Similar to previous demonstrations, a dataset of 6 inputs ($X_1\ \mathrm{to}\ X_6)$ and 1 output ($Y$) with 2000 data points were provided by the first researcher. The output function was unitless but was known to be the cotangent of an angle, one input variable was an angle ($X_1$), one input variable was a normalized speed ($X_2)$, and all other input variables were dummy variables which were immediately removed from the study. 

This resulted in a number of constraints to the function generator, allowing to generate a small number of base functions as listed in Table~\ref{Table_5}. 
\begin{table}[!htb]
\setlength\extrarowheight{10pt}
\caption{List of base functions, \textit{Z}, considered in demonstration 3}
\centering
\begin{tabular}{lll}
\toprule
	Category & Initial Base Functions (\textit{Z}) \\ \hline 
	Cross/Power & $X^a_2$   \\ 
	Trigonometric & $sin(cX_1$) , $cos(cX_1$) , $tan({cX}_1$) ,$\ cot({cX}_1$) , $sec(cX_1$) ,$\ csc(cX_1$) \\
\bottomrule
\end{tabular}
\[a\in \left[-4,\ -2,-1,-0.5,\ 0.5,\ 1,\ 2,\ 4\right], c\in \left[\ 1,\ 2,\ 5\right]\]
\label{Table_5}
\end{table}
Returning to the original research concept in this paper, a neural network was trained repeatedly with combinations of candidate variables. In each instance, the neural network was trained using one function of $X_1$ and one function of $X_2$. The models were ranked according to their RMSEs. Considering the 40 models that resulted in predictions with the lowest error, a set of six functions (two of the angle variable and four of the speed variable) were identified as most likely contributing to the efficacy of the neural network:
\begin{equation} \label{EQ_17} 
tan(X_1), csc(X_1), X^{-4}_2,\ X^{-2}_2, X^{-1}_2, X^{-0.5}_2 
\end{equation} 
These six functions were then compared with the mathematical expression used to generate the data for demonstration three, which is an expression of the angle of deflection of supersonic air of a given velocity affected by a shockwave at a given shock angle \cite{potter2016mechanics}:
\begin{equation} \label{EQ_18} 
{\mathrm{cot} \theta \ }={\mathrm{tan} \beta \ }\left[\frac{\left(\gamma +1\right)M^2}{2\left(M^2{sin}^2\beta -1\right)}-1\right] 
\end{equation} 
For which ${M=X}_2$ is the Mach number, $\beta =X_1$ is the oblique shock angle, $\theta =Y$ is the deflection angle and $\gamma $ is a constant (heat capacity ratio). At first glance, the results were unsatisfying; only one of the six functions was explicitly used in the equation, the other five were not, and one of the functions of the input variables used repeatedly in the equation was not captured in the set of six. However it was noted that the mathematical expression used to generate the data could be manipulated in a number of ways, through a process we generally call ``algebra''. For example, it turned out that the mathematical expression in Equation \ref{EQ_18} could be equivalently written using three of the six identified functions:
\begin{equation} \label{EQ_19} 
{\mathrm{cot} \theta \ }={\mathrm{tan} \beta \ }\left[\frac{{\left(\gamma +1\right)}/{2}}{{\left({\mathrm{csc} \beta \ }\right)}^{-2}-M^{-2}}-1\right] 
\end{equation} 
Further, given the capability of a neural network to determine power functions of a variable, it was not surprising that the other three functions also showed signals. After all, 
\begin{equation} \label{EQ_20} 
X^{-2}_2={\left(X^{-4}_2\right)}^{0.5}={\left(X^{-1}_2\right)}^2 
\end{equation} 
This leads to a bit of a conundrum. A neural network can be used to determine which functions may be directly included in the expression used to generate the data, but there can be many such functions given that there can be many equivalent expressions that precisely describe the same phenomenon.

\section{Discussion}
 In this study, one researcher used governing equations for three different physical phenomena to create datasets of six inputs and one output in each case. These were equation of motion (demonstration 1), equation for \textit{Nusselt }number over a circular cylinder in convective heat transfer (demonstration 2), and angle of deflection in oblique shock waves (demonstration 3). In a blind study, the second researcher attempted different iterative machine learning methods to extract the underlying expressions. Summary of the results and methods are listed under Table~\ref{Table_6}. 
\begin{table}[!htb]
\setlength\extrarowheight{10pt}
\caption{Summary of three demonstration in this study}
\centering
\begin{tabular}{|c|c|p{3cm}|}
\toprule
	Underlying Physical Equation & Obtained Expressions & Method \\ \hline
	$x=x_0+v_0t+\frac{1}{2}at^2$  & $x_0+v_0t+\frac{1}{2}at^2$ & Iterative Correlation Analysis and NN training \\ 
	$Nu=0.3+\frac{0.62{Re}^{{1}/{2}}{Pr}^{{1}/{3}}}{{\left[1+{\left({0.4}/{Pr}\right)}^{{2}/{3}}\right]}^{{1}/{4}}}$${\left[1+{\left(\frac{Re}{282,000}\right)}^{{5}/{8}}\right]}^{{4}/{5}}$ & $\frac{Re{Pr}^{{1}/{2}}\sqrt{\frac{\mu }{\rho }}}{{\left(0.374+{Re}^{{1}/{2}}\sqrt{\frac{\mu }{\rho }}\right)}^2}$ & Iterative Correlation Analysis \\ 
	${\mathrm{cot} \theta \ }={\mathrm{tan} \beta \ }\left[\frac{{\left(\gamma +1\right)}/{2}}{{\left({\mathrm{csc} \beta \ }\right)}^{-2}-M^{-2}}-1\right]$ & \pbox{20cm}{$tan(\beta $), $csc(\beta $) \\ $M^{-4}$,$\ M^{-2}$, $M^{-1}$, $M^{-0.5}$} & Iterative NN training \\ 
\bottomrule
\end{tabular}
\label{Table_6}
\end{table}
One observation from these demonstrations was that for each problem, a certain variation of the approach was more effective to uncover underlying expressions. For demonstration 1, a combination of correlation analysis and NN training was used. This demonstration showed that by expanding the list of base functions for correlation analysis (i.e. Table~\ref{Table_2}), better results are obtained. For demonstration 2, only the correlation analysis with expanded base functions were used. The final expression had several components similar to the ones in the closed-form solution. However, there were some differences. This demonstration showed that the knowledge of input units can be effective in guiding the proposed approach. For demonstration 3, units of the inputs were provided to the second researcher. This reduced the number of base functions to the point that NN training could be performed with many combinations of base functions. Although several underlying expressions were detected, some of them were only derivations of exact expressions in the closed-form solution.

\section{Summary and Future Work}
In summary, we proposed an iterative machine learning approach enriched with common functional forms of physical phenomena for discovering theories and underlying expressions. This showed to be a promising approach in some cases leading to uncovering explicit closed-form solutions. For some cases, however, challenges were observed including limitation on the number of base functions, and false positives for derivations of exact expressions. However, the proposed method can be effective as an starting point to analyze physical phenomena and extract underlying expressions. 

Aside from feature transformation, we would like to consider other established SciML approaches in the future. Specifically, we would like to consider the effects of optimization approach, activation function and loss function enriched with domain knowledge. For example, for heat transfer problems, exponential-based activation functions correlate better with the physics of the problem. Moreover, we would like to consider the effect of NN architecture in terms of both hidden layers and neurons on the sensitivity of results on base functions. This can be done in a systematic way to observed changes in correlation of base functions to output. Finally, we would like to employ data k-partitioning to analyze the effect of based functions to train within one region and predict trends in other data regions.

\bibliographystyle{unsrt}  
\bibliography{paper}

\end{document}